\documentclass{llncs}
\usepackage{graphicx}
\begin{document}
\title{Bidirectional Attention for SQL Generation}
\author{
Tong Guo\inst{1}
Huilin Gao\inst{2}
}
%

\institute{Lenovo AI Lab\and
China Electronic Technology Group Corporation Information Science Academy, Beijing, China}
\maketitle              % typeset the header of the contribution
\begin{abstract}
Generating structural query language (SQL) queries from natural language is a long-standing open problem. Answering a natural language question about a database table requires modeling complex interactions between the columns of the table and the question. In this paper, we apply the synthesizing approach to solve this problem. Based on the structure of SQL queries, we break down the model to three sub-modules and design specific deep neural networks for each of them. Taking inspiration from the similar machine reading task, we employ the bidirectional attention mechanisms and character-level embedding with convolutional neural networks (CNNs) to improve the result. Experimental evaluations show that our model achieves the state-of-the-art results in WikiSQL dataset.

\end{abstract}
\section{Introduction}
In recent years, with the explosive development of deep learning techniques~\cite{ref_proc1}, the problem of generating SQL from natural language has been attracting considerable interest recently. We refer to this problem as the natural-language-to-SQL problem (NL2SQL). Relational databases store the structured data, which is a large amount of entities or numbers. Due to the large number of entities or numbers, which enlarge the word vocabulary for deep learning for natural language processing. And the larger vocabulary will make it harder to find the exact data for a natural language query. For example, in question-answering (QA) problem, we use matching network~\cite{ref_arXiv preprint arxiv1} to get the best answer given a question. In the QA problem, the larger amount of entities will lead to the larger class number of answer and will decrease the accuracy of other deep learning models. But generation of SQL from natural language is a good way to handle this application, which could leverage the benefit of relational database itself. 

The study of translating natural language into SQL queries has a long history. Recent works consider deep learning as the main technique. ~\cite{ref_arXiv preprint arxiv2} employs an improved encoder-decoder framework based on neural machine translation
\cite{ref_proc2,ref_arXiv preprint arxiv3} to solve this problem. ~\cite{ref_proc3} uses augmented pointer network~\cite{ref_proc4} to tackle this task, which is a attentional sequence to sequence model as sequence neural semantic parser and achieves state-of-the-art results on a variety of semantic parsing datasets. In~\cite{ref_arXiv preprint arxiv4}, Seq2SQL breaks down this task to several sub-modules or sub-SQL to solve and incorporates execution rewards in reinforcement learning. But Seq2SQL has the "order-matter" problem, which means the order of the two conditions in the WHERE clause does not affect the execution results of the query. It is well-known that the order   affects the performance of a sequence-to-sequence style model~\cite{ref_proc5}. SQLNet~\cite{ref_arXiv preprint arxiv5} introduces the attention mechanisms~\cite{ref_proc6} for the model of~\cite{ref_arXiv preprint arxiv4} and solve the "order-matter" problem by proposing the sequence-to-set technique for the WHERE clause of SQL.

The problem of NL2SQL can be considered as a special instance to the more generic semantic parsing problem. There are many works considering parsing natural language into a logical form
\cite{ref_arXiv preprint arxiv6,ref_proc7,ref_proc8,ref_proc9,ref_proc10,ref_article1,ref_proc11,ref_arXiv preprint arxiv7}. Parsing natural language into a logical form has a wide application in question answering and dialog system. And there are some datasets such as GeoQuery~\cite{ref_proc12} and ATIS~\cite{ref_proc13}.

Our main contributions in this work are three-fold. First, we apply bidirectional attention to add the relevant information between two sequences for prediction. Second, we leverage character-level embedding and convolutional neural networks (CNNs) to maps each word to a vector space as the new embedding input.  Last, the model we design achieves the state-of-the-art on the WikiSQL dataset.
The code is available.
\footnote[1]{\url{https://github.com/guotong1988/NL2SQL}}

\section{Task Description}
In the NL2SQL task, given a question and a database table, the machine needs to generate a SQL to query the database table, and find the answer to the question. The question is described as a sequence of word tokens: 
$Q = \{w_1,w_2,...,w_n\}$
where $n$ is the number of words in the question, and the table is described as a sequence of columns $C=\{c_1,c_2,...,c_n\}$ , where $m$ is the number of columns in the table. The table also has a number of data rows which contains the answer to the question.

We now explain the WikiSQL dataset~\cite{ref_arXiv preprint arxiv4}, a dataset of 80654 hand-annotated examples of questions and SQL queries distributed across 24241 tables from Wikipedia. We present an example in Fig. 1. It has several features:

(1) Our model synthesizes a SQL query under the condi-tion that the table which the SQL will execute is determined. In other words, our model need not predict the exact table in the all tables.

(2) The SQL has a fixed structure: SELECT \$COLUMN1 [\$AGG] FROM TABLE WHERE \$COLUMN2 EQUALS \$VALUE1 [AND \$COLUMN3 EQUALS \$VALUE2], where there are one column or one aggregator in the SELECT clause and there are 0-4 conditions in the WHERE clause. Although there is no JOIN clause in the SQL, the task is still challenging as the baseline achieves.

\begin{figure}
\centering
\includegraphics[width=\textwidth]{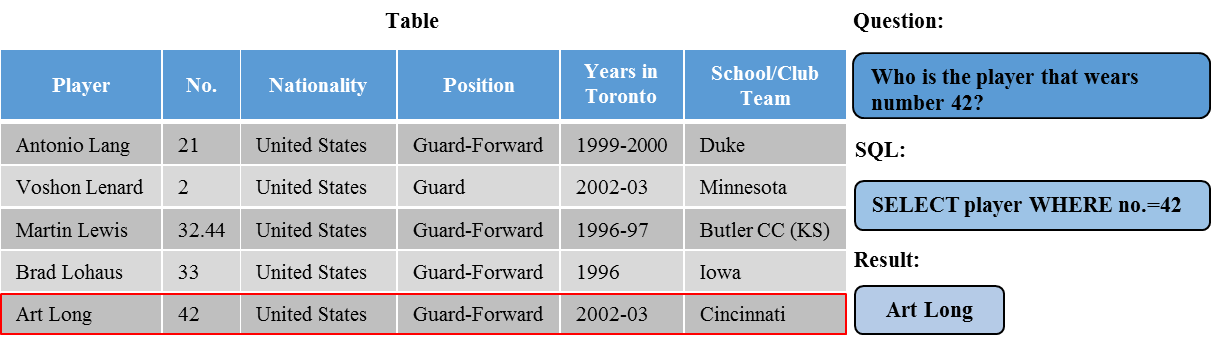}
\caption{An example of the WikiSQL task.} \label{fig1}
\end{figure}

\section{Model}
We present the overall solution for this problem in Fig. 2.
Before we describe the main SQL generation part of our model, we first describe the bi-directional attention mechanism~\cite{ref_proc14} for two sequences.

\begin{figure}
\centering
\includegraphics[width=\textwidth]{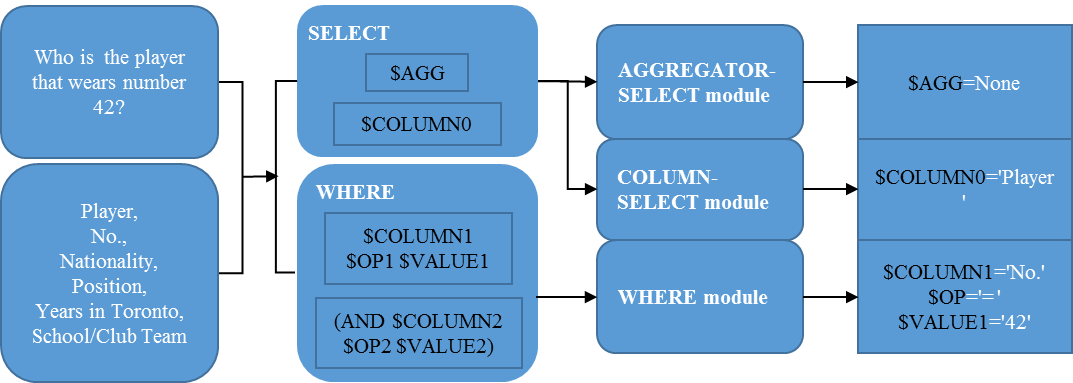}
\caption{The overall problem and solution description.} \label{fig2}
\end{figure}

The reason and inspiration to use bi-directional attention is from the machine comprehension task. In the SQuAD~\cite{ref_arXiv preprint arxiv8} machine comprehension task, we input the paragraph and question to the model and find the answer string in the paragraph. And in the SQL generation task, we input the question and columns and find the answer column in the columns. The two tasks are very similar in this perspective.
\subsection{Bi-attention}
Bi-attention is an extended form of attention mechanism.  The attention mechanism is the information of the most relevant words of second sequence to each word of first sequence. The bi-attention also computes the information signifies which words of first sequence have the closest similarity to one of the words of second sequence.
\subsubsection{Forward Attention}
Suppose we have two sequence of vector representation, $S_1$ and $S_2$ , which dimension is $R^{k_1 \times d}$  and $R^{k_2 \times d}$ , where $d$ is the features dimension size.
Then we compute the co-attention matrix  $M \in R^{k_1 \times k_2}$:
\begin{equation}
M=S_1 \times W  \times {S_2}^T
\end{equation}
where $W \in R^{h \times h}$  is a trainable matrix. Eq. 1 contains the similarities information of each token between the two sequences. Then we apply softmax operation to the second dimension of the matrix $M$:
\begin{equation}
M_1=softmax(M)
\end{equation}
Then we reshape $M_1$ to $M_2 \in R^{k_1 \times k_2 \times 1}$ and reshape $S_1$ to $S_{11} \in R^{k_1 \times 1 \times d}$  and apply element-wise multiplication to get $M_3 \in R^{k_1 \times k_2 \times d}$:
\begin{equation}
M_3=M_2 \times S_{11}
\end{equation}
Note that the  $\cdot$  operation contains the broadcasting mechanism of NumPy. Then we reduce the sum of the second dimension of $M_2$ to get the representation of forward attention information $A_1 \in R^{k_1 \times d}$:
\begin{equation}
A_1=sum(M_3)
\end{equation}

\subsubsection{Backward Attention}
Suppose we already have the co-attention matrix $M \in R^{k_1 \times k_2}$  in Eq. 1. Then we reduce the max value of the first dimension of $M$:
\begin{equation}
M_3=max(M)
\end{equation}
where $M_3 \in R^{1 \times k_2}$ Then we apply softmax to $M_3$:
\begin{equation}
M_4=softmax(M_3)
\end{equation}
Then we reshape $M_4$ to $M_5 \in R^{k_2 \times 1}$ and apply element-wise multiplication with broadcasting mechanism:
\begin{equation}
M_6=M_5 \times S_2
\end{equation}
where $M_6 \in R^{k_2 \times d}$ Then we reduce the sum of the first dimension of $M_6$ to get $M_7 \in R^{1 \times d}$:
\begin{equation}
M_7=sum(M_6)
\end{equation}
Then we compute the element-wise multiplication $A_2=M_7 \times S_1$  to get the representation of backward attention information $A_2 \in R^{k_1 \times d}$. Note that the dimension of back-ward attention representation and forward attention representation are equal and are the same as the sequence $S_1$ dimension. In the next section we use the bi-attention mechanism for several components of our model.

\subsection{Our Model}
In this section, we present our model to tackle the WikiSQL task. As shown in Fig. 3, our model contains four modules:

\begin{figure}
\centering
\includegraphics[width=\textwidth]{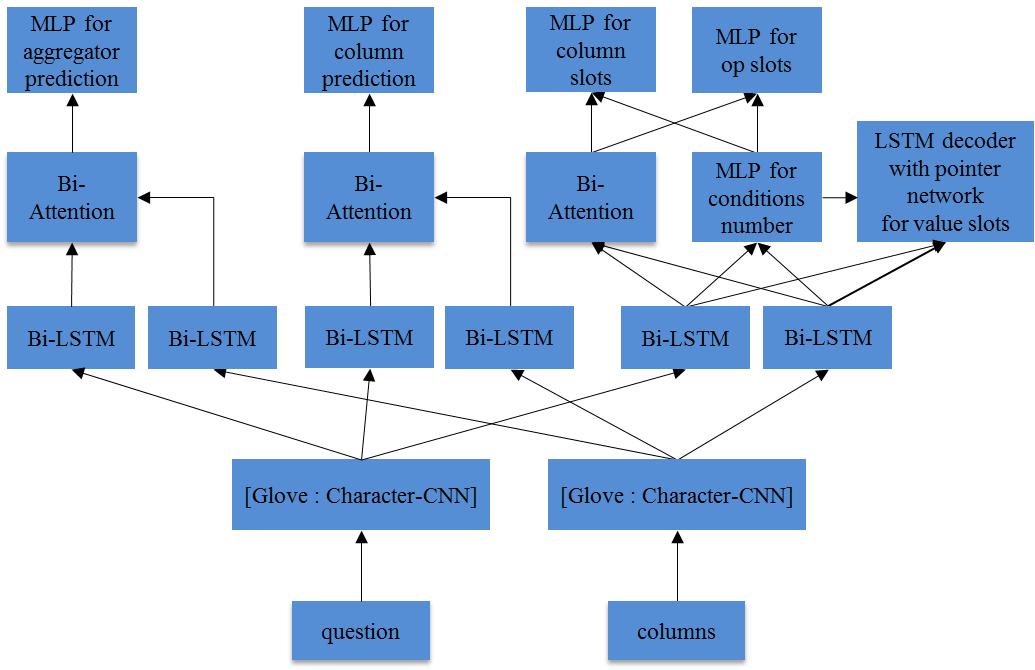}
\caption{Overall architecture of our model.} \label{fig3}
\end{figure}

(1) The character-level and word-level embedding layer to map each character and word to vector space. The embedding layer is shared by the next three modules. (2) The COLUMN-SELECT module which predict the column of SELECT clause. (3) The AGGREGATOR-SELECT module which predict the aggregator of SELECT clause. (4) The WHERE module which predict the conditions of WHERE clause.

The detailed description of our model is provided as follows.

\subsubsection{Character-level embedding and word-level embedding}
We use the character-level GloVe~\cite{ref_proc15} pre-trained 300 dimension to initialize the character-level embedding $E_c \in R^{q \times w \times d}$, where $q$ is the word number and $w$ is the character number of each word and $d$ is 300. We leverage convolutional neural networks to get the next representation of $E_c$. We use three convolution kernels, which sizes are height 1 * width 5, height 1 * width 4, height 1 * width 3. The convolution layer and max-pooling layer are 1 as~\cite{ref_proc16} did. The input channel is 100 and output channel is 100 so the last dimension of 3 convolution results can concatenate to 300. After the max pooling and concatenation, the dimension of the final result is $q \times d$, which dimension is the same as the word embedding dimension.

We use the word-level GloVe pre-trained with 300 size to initialize the word-level embedding $E_w \in R^{q \times d}$. As for the words which are not in the GloVe, we initialize them to 0. The experiment shows that if we initialize the words which are not in GloVe to a random value and make them trainable, the result decreases. 

As one column contains several words, we encode the words of one column into one vector representation by running after a LSTM~\cite{ref_article2}. We take the last state of the LSTM for the representation of one column and consider it as one item of columns, which is the same as one word in the question.

\subsubsection{COLUMN-SELECT module}
In this sub-task, the inputs are questions and column names, the outputs are one column in the column names. So we need to
capture the attention info of questions and column names and then leverage the attention info to output a prediction over the column names. We did it as follows.

Each token is represented as a one-hot vector and fed into a word embedding vector before feeding them into the bi-directional LSTM. We have the question embedding $E_Q \in R^{q \times d}$ and the column embedding $E_{col} \in R^{c \times d}$, where $q$ is the max word number of questions and $c$ is the columns number of a table. The embedding $E_Q$ and $E_{col}$ can be computed as the hidden states of a bi-directional LSTM  respectively and get the bi-LSTM encoded representation $H_Q \in R^{q \times h}$ and $H_{col} \in R^{c \times h}$, where $h$/2 is the hidden size of LSTM.

Then we apply bi-directional attention to $H_Q$ and $H_{col}$ according to Eq. 1 to Eq. 8, where $H_{col}$ is the first sequence $S_1$ and $H_Q$ is the second sequence $S_2$, to get the bi-attention info $B_f \in R^{c \times h}$ and $B_b \in R^{c \times h}$. Then we concatenate the last dimension of forward attention info $B_{f-sel}$ and the backward attention info $B_{b-sel}$ to  $[B_{f-sel}:B_{b-sel}] \in R^{c \times 2h}$ and apply the operations below to get the final prediction $P_{sel} \in R^c$ for column in the SELECT clause:
\begin{equation}
P_{sel}=softmax(W_3tanh(W_1[B_{f-sel}:B_{b-sel}]+W_2H_c))
\end{equation}
where $W_1 \in R^{2h \times h}$, $W_2 \in R^{h \times h}$ and $W_3 \in R^{h \times 1}$ are all trainable weights and $P_{sel}$ is the probability distribution over all columns of one table.

\subsubsection{AGGREGATOR-SELECT module}
There are 5 types of aggregation keywords in SQL: 'MAX', 'MIN', 'COUNT', 'SUM', 'AVG'. The experiment of SQLNet shows that the column name input do not impact the prediction result. So we only need to input the question and predict the class of aggregation keywords. So we consider this sub-task as a text classification problem. Then we have the question embedding $E_Q \in R^{q \times d}$ and the Bi-LSTM encoded representation $H_Q \in R^{q \times h}$. Then we compute the final prediction for aggregator $P_{agg} \in R^6$:  
\begin{equation}
P_{agg}=softmax(W_2tanh(W_1H_Q))
\end{equation}
where $W_1 \in R^{h \times h}$ and $W_2 \in R^{h \times 6}$ are all trainable weights and sum apply to the first dimension and $P_{agg}$ is the probability distribution over 6 choices of SQL aggregators.

\subsubsection{WHERE module}
The WHERE clause is the most challenging part. The order of conditions does not matter, so we predict the probability of column slots and choose the top columns as a set. We predict the number of conditions and the column slots first. Then we leverage the column predictions to choose from the columns candidates. Then we use the chosen columns as embedding input to predict the operator slots and value slots for each column. We describe them below.

\paragraph{(1) Condition number}
Suppose the number of conditions of WikiSQL is ranging from 0 to $N$ and we consider it as a ($N$+1)-class classification problem. We compute the bi-LSTM hidden states $H_Q$ of the question embedding $E_Q$ and compute the number of conditions   as below:
\begin{equation}
K=argmax(softmax(W_2tanh(W_1H_Q)))
\end{equation}
where $W_1 \in R^{h \times h}$ and $W_2 \in R^{h \times N}$ are all trainable weights.

\paragraph{(2) Column slots}
In this sub-task, taking questions and column names as input,
we need to predict the top $K$ column names. So we leverage the bi-attention info of the questions and column names to output the probability over the column names. We compute bi-attention information $[B_f:B_b] \in R^{c \times 2h}$ of the question hidden states $H_Q \in R^{q \times h}$ and column hidden states $H_{col} \in R^{c \times h}$. Then we compute the bi-attention information according to Eq. 1 to Eq. 8 and compute final column prediction $P_{col} \in R^c$, which is the same computation as COLUMN-SELECT module. We choose the top $K$ scores of $P_col$ as the prediction of $K$ columns and pad 0 to $N$ columns. We leverage the chosen and padded LSTM representation $H_{topcol} \in R^{N \times h}$ for the predictions of operator slots and value slots.

\paragraph{(3) Operator slots}
There are 3 type of operator keywords in SQL: 'GT', 'LT', 'EQL', which indicates 'greater than', 'less than', 'equal' separately. We start from the hidden states $H_Q \in R^{q \times h}$ and $H_{col} \in R^{c \times h}$. And we use the result of column name slots predictions $P_{col}$ to choose the top $K$ column from $H_{col}$ and get $H_{topcol} \in R^{N \times h}$. Then we apply the bi-attention of Eq. 1 to Eq. 8 to get the final attention representation $[B_{f-op}:B_{b-op}] \in R^{N \times 2h}$ for $H_Q$ and $H_{topcol}$, which is the concatenation of the last dimension of the two sequence representation. Then the computation is to get predictions $P_{op} \in R^{N \times 3}$ for each condition:
\begin{equation}
P_{op}=softmax(W_3tanh(W_1([B_{f-op}:B_{b-op}]+W_2H_{topcol}))
\end{equation}
where $W_1 \in R^{2h \times h}$, $W_2 \in R^{h \times h}$ and $W_3 \in R^{h \times 3}$ are all trainable weights and $P_{op} \in R^{N \times 3}$ is the probability distribution over 4 choices of condition operator for each column. 

\paragraph{(4) Value slots}
As one value slot must corresponding to one column slot and we need to leverage the predicted columns info, so we employ the sequence-to-sequence structure to generate the values by taking the predicted columns info as input. The structure is well-developed: suppose we have an input sequence, and we employ an encoder to encode the input sequence into a vector. Then we employ a decoder to decode the output sequence from the vector.

We employ bi-LSTM to be the encoder which take the question embedding $E_Q \in R^{q \times d}$ and predicted column info  as input and the encoder’s output is $H_Q \in R^{q \times h}$. At decoder phase we need to predict the value which is a sub-string of the question, so we use pointer network~\cite{ref_proc4} to point to the sub-string of question. The LSTM decoder of pointer network is unidirec-tional and 2 layers. In training, the LSTM decoder takes the ground truth $G_i \in R^{N \times q \times m}$ as input and outputs the $G_o \in R^{N \times q \times h}$, where $m$ is the max word number and is one-hot encoding. Then $H_Q \in R^{q \times h}$ and $H_{topcol} \in R^{N \times h}$ participate the next computation:
\begin{equation}
P_{val}=softmax(W_4tanh(W_1G_o+W_2H_q+W_3H_{topcol}))
\end{equation}
Where the inputs $G_o$, $H_Q$ and $H_{topcol}$ are expanded to the same dimension and $W_1 \in R^{h \times h}$, $W_2 \in R^{h \times h}$, $W_3 \in R^{h \times h}$ and $W_4 \in R^{h \times 1}$ are all separated trainable weights. The output of the pointer network is $P_{val} \in R^{N \times p}$, where $q$ is the question length. In engineering we flat the specific dimension for the computation. For example, suppose we have batch size dimension $B$ and $N$ conditions as the second dimension, then we flat the dimension to $B$ * $N$ as the first dimension. Note that we generate the condition values for each of the $K$ conditions. The END token also appears in the question and the model stops generating for this slot when the END token is predicted. We prepare the exact ground truth for each sub-module of WHERE module and give each sub-module of WHERE module a separated loss.

\subsubsection{Loss function }
We use the cross-entropy loss for the prediction of COLUMN-SELECT module and AGGREGATOR-SELECT module. As for the WHERE module, we also use cross-entropy loss for the value slots and operator slots.

As the prediction for the columns in the WHERE module is a target set, we need to penalize the predicted columns that are not in the ground truth. So we design the loss function for the prediction of columns set:
\begin{equation}
L_{wherecol}=-\sum_{j=1}^K(\gamma C_jlog(P_{col-j})+(1-C_j)log(1-P_{col-j}))
\end{equation}
where we choose $\gamma$=3 and $C_j$=1 if the ground truth contains the $j$-column, $C_j$=0 if the ground truth does not contain the $j$-column. The final objective function is：
\begin{equation}
L=L_{agg}+L_{sel}+L_{whe}
\end{equation}
where $L_{agg}$, $L_{sel}$, $L_{whe}$ are the loss of AGGREGATOR-SELECT, COLUMN-SELECT and WHERE module separately. 

\section{Experiments}
In this section, we present more details of the model and the evaluation on the dataset. We also analyze the evaluation result.

\subsection{Experimental Setting}
We tokenize the sentences using Stanford CoreNLP~\cite{ref_proc17}. The LSTM contains 2 layers and the size of LSTM hidden states h is 50 and the output size of bi-LSTM is 100. The dropout~\cite{ref_arXiv preprint arxiv9} for LSTM cell is 0.3. We use different LSTM weights for predicting different slots. Each LSTM which encodes the embedding is an independent weight. Although we do not share the bi-LSTM weight, we find that sharing the same word em-bedding vector is better. Therefore, different components in our model only share the word embedding. We use the Adam optimizer~\cite{ref_proc18} with learning rate 0.001 and 0 weight decay to minimize the loss of Eq. 15. We train the model for 100 epochs with fixed word embedding and trainable character embedding. Then we use the pre-trained 100 epoch model to train the next 100 epoch with all traina-ble embeddings. The character-level embedding are all trainable in 0 to 200 epoch. The batch size is 64. We ran-domly re-shuffle the training data in each epoch. In addition, our final model is chosen as the models that perform the best on development set in each part in the process of training. We implement all models using PyTorch~\cite{ref_url1}.

\subsection{Evaluation}
We evaluate our model on the WikiSQL dataset. The decomposition results are presented in Tab. 1 and the overall results are presented in Tab. 2. We display the separated results of each module and the query-match accuracy which compare whether two SQL queries match exactly. From the evaluation result we find that bi-attention mechanisms mainly improve the WHERE clause result and character-level embedding mainly improve the COLUMN-SELECT clause. The execution result is higher because different SQL may obtains the same result. For example, the two SQL queries SELECT COUNT (player) WHERE No. = 23 and SELECT COUNT (No.) WHERE No. = 23 produce the same result in the table of Fig. 1.

\begin{table}
\caption{Our baselines are Seq2SQL~\cite{ref_arXiv preprint arxiv4} and SQLNet~\cite{ref_arXiv preprint arxiv5}. The third row is our model with bi-attention and +char-emb means our model with CNN-based character-level embedding. $Acc_{agg}$ and $Acc_{sel}$ indicate the accuracy on the aggregator and column prediction accuracy on the SELECT clause, and $Acc_{where}$ indicates the accuracy to generate the WHERE clause.}\label{tab1}
\centering
\begin{tabular}{|l|l|l|l|l|l|l|}
\hline
& \multicolumn {3} {|c|} {dev} & \multicolumn {3} {|c|} {test}\\
\cline{2-7} 
& $Acc_{agg}$ & $Acc_{sel}$ & $Acc_{where}$ & $Acc_{agg}$ & $Acc_{sel}$ & $Acc_{where}$\\
\hline
Seq2SQL & 90.0$\%$ & 89.6$\%$ & 62.1$\%$ & 90.1$\%$ & 88.9$\%$ & 60.2$\%$\\
\hline
SQLNet & 90.1$\%$ & 91.5$\%$ & 74.1$\%$ & 90.3$\%$ & 90.9$\%$ & 71.9$\%$\\
\hline
Our Model & 90.1$\%$ & 91.1$\%$ & 74.6$\%$ & 90.3$\%$ & 90.8$\%$ & 72.8$\%$\\
\hline
+char-emb & 90.1$\%$ & 92.5$\%$ & 74.7$\%$ & 90.3$\%$ & 91.9$\%$ & 72.8$\%$\\
\hline
\end{tabular}
\end{table}

\begin{table}
\caption{ Overall result on the WikiSQL task. The "result match" indicates the execution of database accuracy and the "query string match" is to compare whether predicted SQL and ground truth SQL match exactly.}\label{tab1}
\centering
\begin{tabular}{|l|l|l|l|l|}
\hline
& \multicolumn {2} {|c|} {dev} & \multicolumn {2} {|c|} {test}\\
\cline{2-5} 
& Result match & Query string match & Result match & Query string match\\
\hline
Seq2SQL & 62.1$\%$ & 53.5$\%$ & 60.4$\%$ & 51.6$\%$\\
\hline
SQLNet & 69.8$\%$ & 63.2$\%$ & 68.0$\%$ & 61.3$\%$\\
\hline
Our Model & 70.3$\%$ & 63.5$\%$ & 68.2$\%$ & 61.5$\%$\\
\hline
+char-emb & 71.1$\%$ & 64.1$\%$ & 69.0$\%$ & 62.5$\%$\\
\hline
\end{tabular}
\end{table}

\subsection{Analysis}
The improvement of COLUMN-SELECT clause which is attributed by CNN-based character-level embedding is around 2$\%$, as the baseline result is already 90$\%$. We think it is because with the help of the character-level embedding, the model can be more robust to the minor difference of a word between training data and test data. The improvement of attention is 2.5$\%$ and the improvement of the bi-attention mechanisms is 3$\%$ to 3.5$\%$. The improvement from attention to bi-attention is 0.5$\%$ to 1$\%$. We also observe that if we initialize the words which are not in the GloVe the random initialization and train the embedding, the result does not improve. The reason is that we do not add the mask technique which set the rare words to a minimal value in the model in order that the rare words do not participate in the activation function such as sigmoid. We consider the mask technique as a future work.

\section{Conclusion}
In this paper, based on the structure of SQL and the observation that a sequence-to-sequence model suffer from the "order-matters" problem, we design specific deep neural network for each sub-string of SQL. In the WHERE prediction module, we choose the top probabilities of the column candidates as the chosen set for the prediction of conditions. We apply the bi-directional attention mechanism and the CNN-based character-level embedding to improve the result. The experimental evaluations show that our model achieves the state-of-the-art results in the WikiSQL dataset.

We observe that the accuracy is around 90$\%$ on the COLUMN-SELECT clause prediction and AGGREGATOR-SELECT clause prediction because the number of candidate column in the SELECT clause is limited to one. The task will be more challenging if the SQL extends to more than one column candidates and more complex cases like ORDER-BY, GROUP-BY or even JOIN. And the technique of NL2SQL can be applied to Knowledge Graph query or other semantic parsing tasks. There will be a lot of work to research.


\begin{thebibliography}{8}
\bibitem{ref_proc1}
A. Krizhevsky, I. Sutskever, and G. Hinton.: Imagenet classification with deep convolutional neural networks. In NIPS (2012)

\bibitem{ref_arXiv preprint arxiv1}
Y.Wu, W. Wu, Z. Li and M. Zhou: Response Selection with Topic Clues for Retrieval-based Chatbots. arXiv preprint arxiv:1605.00090 (2016)

\bibitem{ref_arXiv preprint arxiv2}
Ruichu Cai, Boyan Xu, Xiaoyan Yang, Zhenjie Zhang, Zijian Li: An Encoder-Decoder Framework Translating Natural Language to Database Queries. arXiv preprint arxiv:1711.06061, Nov 2017

\bibitem{ref_proc2}
Casacuberta F, Vidal E.: Machine Translation using Neural Networks and Finite-State Models. In Proceedings of the 7th International Conference on Theoretical and Methodological Issues in Machine Translation (1997)

\bibitem{ref_arXiv preprint arxiv3}
Cho, K.; Van Merrienboer, B.; Gulcehre, ¨ C.; Bahdanau, D.; Bougares, F.; Schwenk, H.; and Bengio, Y.: Learning phrase representations using rnn encoder-decoder for statistical machine translation. arXiv preprint arXiv:1406.1078 (2014)

\bibitem{ref_proc3}
Li Dong and Mirella Lapata: Language to logical form with neural attention. Meeting of the Association for Computational Linguistics, January 2016

\bibitem{ref_proc4}
Oriol Vinyals, Meire Fortunato, and Navdeep Jaitly: Pointer networks. International Conference on Neural Information Processing Systems. MIT Press (2015)

\bibitem{ref_arXiv preprint arxiv4}
Victor Zhong, C. Xiong, and R. Socher. Seq2SQL: Generating Structured Queries from Natural Language using Reinforcement Learning. arXiv preprint arxiv:1709.00103, Nov 2017

\bibitem{ref_proc5}
Oriol Vinyals, Samy Bengio, and Manjunath Kudlur: Order matters: Sequence to sequence for sets. In ICLR (2016)

\bibitem{ref_arXiv preprint arxiv5}
Xiaojun Xu, Chang Liu and Dawn Song. SQLNet: Generating Structured Queries From Natural Language Without Reinforcement Learning. arXiv preprint arXiv:1711.04436, Nov 2017

\bibitem{ref_proc6}
Dzmitry Bahdanau, Kyunghyun Cho, and Yoshua Bengio: Neural machine translation by jointly learning to align and translate. ICLR (2015)

\bibitem{ref_arXiv preprint arxiv6}
Luke S Zettlemoyer and Michael Collins: Learning to map sentences to logical form: Structured classification with probabilistic categorial grammars. arXiv preprint arXiv:1207.1420 (2012)

\bibitem{ref_proc7}
Yoav Artzi and Luke Zettlemoyer: Bootstrapping semantic parsers from conversations. In Proceedings of the conference on empirical methods in natural language processing, pp. 421–432. Association for Computational Lin-guistics (2011)

\bibitem{ref_proc8}
Yoav Artzi and Luke Zettlemoyer: Weakly supervised learning of semantic parsers for mapping instructions to actions. Transactions of the Association for Computational Linguistics, 1:49–62 (2013)

\bibitem{ref_proc9}
Qingqing Cai and Alexander Yates: Large-scale semantic parsing via schema matching and lexicon extension. In ACL (2013)

\bibitem{ref_proc10}
Siva Reddy, Mirella Lapata, and Mark Steedman: Large-scale semantic parsing without questionanswer pairs. Transactions of the Association for Computational Linguistics, 2:377–392 (2014)

\bibitem{ref_article1}
Percy Liang, Michael I Jordan, and Dan Klein: Learning dependency-based compositional semantics. Computational Linguistics, 39(2):389-446 (2011)

\bibitem{ref_proc11}
Chris Quirk, Raymond J Mooney, and Michel Galley: Language to code: Learning semantic parsers for if-this-then-that recipes. In ACL (1), pp. 878–888 (2015)

\bibitem{ref_arXiv preprint arxiv7}
Xinyun Chen, Chang Liu, Richard Shin, Dawn Song, and Mingcheng Chen: Latent attention for if-then program syn-thesis. arXiv preprint arXiv:1611.01867 (2016)

\bibitem{ref_proc12}
Lappoon R. Tang and Raymond J. Mooney: Using multiple clause constructors in inductive logic programming for semantic parsing. In Proceedings of the 12th European Conference on Machine Learning, pp. 466–477, Freiburg, Germany (2001)

\bibitem{ref_proc13}
P. J. Price: Evaluation of spoken language systems: the atis domain. In Speech and Natural Language: Proceedings of a Workshop Held at Hidden Valley, Pennsylvania, June 24-27, pp. 91–95 (1990)

\bibitem{ref_proc14}
M. Seo, A. Kembhavi, A. Farhadi, et al: Bidirectional Attention Flow for Machine Comprehension. In Proceedings of ICLR (2017)

\bibitem{ref_arXiv preprint arxiv8}
Rajpurkar P, Zhang J, Lopyrev K, et al: Squad: 100,000+ questions for machine comprehension of text[J]. arXiv pre-print arXiv:1606.05250 (2016)

\bibitem{ref_proc15}
Jeffrey Pennington, Richard Socher, and Christopher Manning. Glove: Global vectors for word representation. In Proceedings of the 2014 conference on empirical methods in natural language processing (EMNLP), pp. 1532–1543 (2014)

\bibitem{ref_proc16}
Yoon Kim: Convolutional neural networks for sentence classification. In EMNLP (2014)

\bibitem{ref_article2}
Sepp Hochreiter and Jurgen Schmidhuber: Long short-term memory. Neural Computation (1997）

\bibitem{ref_proc17}
Christopher D. Manning, Mihai Surdeanu, John Bauer, Jenny Finkel, Steven J. Bethard, and David McClosky: The Stanford CoreNLP natural language processing toolkit. In Association for Computational Linguistics (ACL) System Demonstrations, pp. 55–60 (2014)

\bibitem{ref_arXiv preprint arxiv9}
G. E. Hinton, N. Srivastava, A. Krizhevsky, I. Sutskever, and R. R. Salakhutdinov: Improving neural networks by preventing coadaptation of feature detectors. arXiv preprint arXiv:1207.0580 (2012)

\bibitem{ref_proc18}
Diederik P. Kingma and Jimmy Ba. Adam: A method for stochastic optimization. In Proceedings of the 3rd International Conference for Learning Representations, San Diego (2015)

\bibitem{ref_url1}
Pytorch. URL http://pytorch.org/ (2017)
\end{thebibliography}
\end{document}